\documentclass{article}
\usepackage[preprint]{spconf,amsmath,graphicx}
\copyrightnotice{\copyright\ IEEE 2019}
\toappear{To appear in {\it Proc.\ ICASSP2019, May 12-17, 2019, Brighton, UK}}
\usepackage{multirow}
\usepackage{booktabs}
\usepackage{enumitem}
\usepackage{subfigure}
\usepackage{makecell}
\usepackage{tablefootnote}

\title{End-to-end Anchored Speech Recognition}
\name{Yiming Wang$^1$\sthanks{This work was done
while the authors were interns at Amazon.}, Xing Fan$^2$, I-Fan Chen$^2$, Yuzong Liu$^2$, Tongfei Chen$^{1*}$, Bj{\"o}rn Hoffmeister$^2$}
\address{
  $^1$ Center for Language and Speech Processing, Johns Hopkins University, Baltimore, MD, USA\\
  $^2$ Amazon.com, Inc., USA\\
\texttt{\{yiming.wang,tongfei\}@jhu.edu},\\ \texttt{\{fanxing,ifanchen,liuyuzon,bjornh\}@amazon.com}
}

\begin{document}
\ninept
\maketitle
\begin{abstract}
Voice-controlled house-hold devices, like Amazon Echo or Google Home, face the problem of performing speech recognition of device-directed speech in the presence of interfering background speech, i.e., background noise and interfering speech from another person or
media device in proximity need to be ignored. We propose two end-to-end models to tackle this problem with information extracted from the \emph{anchored segment}. The anchored segment refers to the wake-up word part of an audio stream, which contains valuable speaker information that can be used to suppress interfering speech and background noise. The first method is called \emph{Multi-source Attention} where the attention mechanism takes both the speaker information and decoder state into consideration. The second method directly learns a frame-level mask on top of the encoder output. We also explore a multi-task learning setup where we use the ground truth of the mask to guide the learner. Given that audio data with interfering speech is rare in our training data set, we also propose a way to synthesize ``noisy'' speech from ``clean'' speech to mitigate the mismatch between training and test data. Our proposed methods show up to 15\% relative reduction in WER for Amazon Alexa live data with interfering background speech without significantly degrading on clean speech.

\end{abstract}
\begin{keywords}
End-to-End ASR, anchored speech recognition, attention-based encoder-decoder network, robust speech recognition
\end{keywords}
\vspace{-2mm}
\section{Introduction}
\vspace{-2mm}
\label{sec:intro}
We tackle the ASR problem in the scenario where a foreground speaker first wakes up a voice-controlled device with an ``anchor word'', and the speech after the anchor word is possibly interfered with background speech from other people or media. Consider the following example:

\quad {\sc Speaker 1}: {\it Alexa, play rock music.}

\quad {\sc Speaker 2}: {\it I need to go grocery shopping.}
 
Here the wake-up word ``Alexa'' is the anchor word, and thus the utterance by speaker 1 is considered as device-directed speech, while the utterance by speaker 2 is the interfering speech. Our goal is to extract information from the anchor word in order to recognize the device-directed speech and ignore the interfering speech. We name this task \emph{anchored speech recognition}. The challenge of this task is to learn a speaker representation from a short segment corresponding to the anchor word. A couple of techniques have been proposed for learning speaker representations, e.g., i-vector~\cite{dehak2011front,saon2013speaker}, mean-variance normalization~\cite{liu1993efficient}, maximum likelihood linear regression (MLLR)~\cite{leggetter1995maximum}. With the recent progress in deep learning, neural networks are used to learn speaker embeddings for speaker verification/recognition~\cite{variani2014deep,mclaren2015advances,snyder2016deep,heigold2016end}. More relevant to our task, two methods---anchor mean subtraction (AMS) and an encoder-decoder network---are proposed to detect desired speech by extracting speaker characteristics from the anchor word~\cite{maas2016anchored}. This work is further extended for acoustic modeling in hybrid ASR systems~\cite{King2017RobustSR}. Speaker-dependent mask estimation is also explored for target speaker extraction ~\cite{delcroix2018single,wang2018deep}.

Recently, much work has been done towards end-to-end approaches for speech recognition~\cite{graves2012sequence,graves2014towards,chorowski2015attention,chan2016listen,chiu2018state,chen2018on}. These approaches typically have a single neural network model to replace previous independently-trained components, namely, acoustic, language, and pronunciation models from hybrid HMM systems. End-to-end models greatly alleviate the complexity of building an ASR system. Compared with the others, the attention-based encoder-decoder models \cite{bahdanau2014neural,chan2016listen} do not assume conditional independence for output labels as CTC based models~\cite{graves2006connectionist,graves2014towards} do. 

We propose two end-to-end models for anchored speech recognition, focusing on the case where each frame is either completely from device-directed speech or completely from interfering speech, but not a mixture of both~\cite{delcroix2018single,chen2018progressive}. They are both based on the attention-based encoder-decoder models. The 
attention mechanism provides an explicit way of aligning each output 
symbol with different input frames, enabling \emph{selective decoding} 
from an audio, i.e., only decode desired speech that is taking place in 
part of the entire audio stream. In the first method, we incorporate the speaker information when calculating the attention energy, which leads to an anchor-aware soft alignment between the decoder state and encoder output. The second method learns a frame-level mask on top
of the encoder, where the mask can optionally be 
learned in the multi-task framework if the gold mask is given. This method will pre-select the encoder output before the attention energy is calculated. Furthermore, since the training data is relatively clean, in the sense that it contains device-directed speech only, we propose a method to synthesize ``noisy'' speech from ``clean'' speech, mitigating the mismatch between training and test data.

We conduct experiments on a training corpus consisting of 1200 hours live data in English from Amazon Echo. The results demonstrate a significant WER relative gain of 12-15\% in test sets with interfering background speech. For a test set that contains only device-directed speech, we see a small relative WER degradation from the proposed method, ranging from 1.5\% to 3\%.

This paper is organized as follows. Section~\ref{sec:model} first gives an overview of the attention-based encoder-decoder model and then presents our two end-to-end anchored ASR models. Section~\ref{sec:synthetic} describes how we synthesize our training data and train our second proposed model in a multi-task fashion. Section~\ref{sec:exp} shows our experiments and results. Section~\ref{sec:conclusion} includes conclusions and future work.

\section{Model Overview}
\label{sec:model}

\subsection{Attention-based Encoder-Decoder Model}
\label{sec:enc_dec}
\begin{figure}
  \centering
  \includegraphics[width=0.4\textwidth]{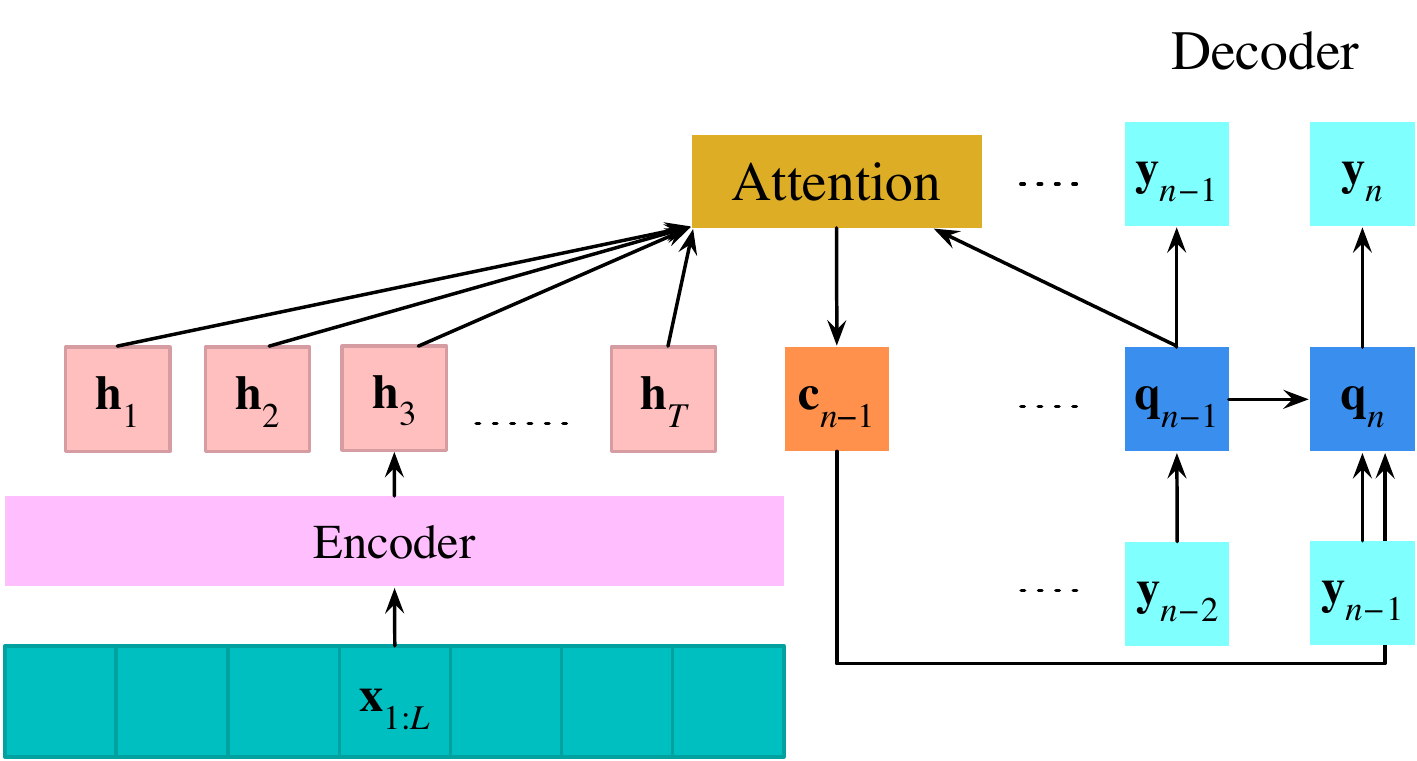}
  \caption{Attention-based Encoder-Decoder Model. It is an illustration in the case of a one-layer decoder. If there are more layers, as in our experiments, an updated context vector $\mathbf{c}_n$ will also be fed into each of the upper layers in the decoder at the time step $n$.}
  \label{fig:enc_dec}
\end{figure}

The basic attention-based encoder-decoder model typically consists of 3 modules as depicted in Fig~\ref{fig:enc_dec}: 1) an encoder transforming a sequence of input features $\mathbf{x}_{1:L}$ into a high-level representation of the features $\mathbf{h}_{1:T}$ through a stack of convolution/recurrent layers, where $T \le L$ due to possible frame down-sampling; 2) an attention module summarizing the output of the encoder $\mathbf{h}_{1:T}$ into a fixed length context vector $\mathbf{c}_{n}$ at each output step for $n \in [1,\ldots,N]$, which determines parts of the sequence $\mathbf{h}_{1:T}$ to be attended in order to predict the output symbol $y_n$; 3) a decoder module taking the context vector $\mathbf{c}_n$ as input and predicting the next symbol $y_n$ given the history of previous symbols $y_{1:n-1}$. The entire model can be formulated as follows:
\begin{eqnarray}
\mathbf{h}_{1:T}&=&\mathrm{Encoder}(\mathbf{x}_{1:L}) \\
\alpha_{n,t}&=&\mathrm{Attention}(\mathbf{q}_n,\mathbf{h}_t) \label{eqn:atten}\\
\mathbf{c}_n&=&\sum_{t} \alpha_{n,t} \mathbf{h}_t \label{eqn:context} \\
\mathbf{q}_n&=&\mathrm{Decoder}(\mathbf{q}_{n-1},[y_{n-1};\mathbf{c}_{n-1}]) \\
y_n&=&\arg\max_v (\mathbf{W}^f \mathbf{q}_n+\mathbf{b}^f)
\end{eqnarray}

Although our proposed methods do not limit itself in any particular attention mechanism,  we choose the 
\emph{Bahdanau Attention}~\cite{bahdanau2014neural} as the attention function for our experiments. So Eqn.~(\ref{eqn:atten}) takes the form of:
\begin{eqnarray}
\omega_{n,t}&=&\mathbf{v}^\top \tanh(\mathbf{W}^q\mathbf{q}_n+\mathbf{W}^h \mathbf{h}_t+\mathbf{b}) \label{eqn:bahdanau}\\
\alpha_{n,t}&=&\mathrm{softmax}(\omega_{n,t}) \label{eqn:atten_weights}
\end{eqnarray}

\subsection{Multi-source Attention Model}
\label{sec:multi-source}
Our first approach is based on the intuition that the attention mechanism should consider both the speaker information and the decoder state -- when computing the attention weights, in addition to conditioning on the decoder state, the speaker information extracted from the input frames is also utilized. In our scenario, the device-directed speech and the anchor word are uttered by the same speaker, while the interfering background speech is from a different speaker. Therefore, the attention mechanism can be augmented by placing more attention probability mass on frames that are more similar to the anchor word in terms of speaker characteristics. 

Formally speaking, besides our previous notations, the anchor word segment is denoted as $\mathbf{w}_{1:L'}$. We add another encoder $\mathrm{S{\text -}Encoder}$ to be applied on both $\mathbf{w}_{1:L'}$ and $\mathbf{x}_{1:L}$ to generate a fixed-length vector $\tilde{\mathbf{w}}$ and a variable length sequence $\mathbf{u}_{1:T}$ respectively:
\begin{eqnarray}
\tilde{\mathbf{w}}&=&\mathrm{Pooling}(\mathrm{S{\text-}Encoder}(\mathbf{w}_{1:L'})) \\
\mathbf{u}_{1:T}&=&\mathrm{S{\text-}Encoder}(\mathbf{x}_{1:L})
\end{eqnarray}
As shown above, $\mathrm{S{\text -}Encoder}$ extracts speaker characteristics from the acoustic features. In our experiments, the pooling function is implemented as \emph{Max-pooling} across all output frames if $\mathrm{S{\text -}Encoder}$ is a convolutional network, or picking the hidden state of the last frame if $\mathrm{S{\text -}Encoder}$ is a recurrent network. Rather than being appended to acoustic feature vector and fed into the decoder\footnote{We tried that in our preliminary experiments but it did not perform well.} as proposed in \cite{King2017RobustSR}, $\tilde{\mathbf{w}}$ is directly involved in computing the attention weights. Specifically, Eqn.~(\ref{eqn:atten_weights}) and Eqn.~(\ref{eqn:context}) are replaced by:
\begin{eqnarray}
\phi_t &=& \mathrm{Similarity}(\mathbf{u}_t,\tilde{\mathbf{w}})\label{eqn:sim1}\\
\alpha_{n,t}^{\textrm{anchor\_aware}} &=& \mathrm{softmax}(\omega_{n,t}+g \cdot \phi_t)\\
\mathbf{c}_n&=&\sum_{t} \alpha_{n,t}^{\textrm{anchor\_aware}} \mathbf{h}_t
\end{eqnarray}
where $g$ is a trainable scalar used to automatically adjust the relative contribution from the speaker acoustic information. $\mathrm{Similarity}(\cdot,\cdot)$ is implemented as dot-product in our experiments. As a result, the attention weights are essentially computed from two different sources: the ASR decoding state, and the confidence of decision on whether each frame belongs to the device-directed speech. We call this model \textit{Multi-source Attention} to reflect the way the attention weights are computed.

\subsection{Mask-based Model}
\label{sec:masked}
The Multi-source Attention model jointly considers speaker characteristic and ASR decoder state when calculating the attention weights. However, since the attention weights are normalized with a  softmax function, whether each frame needs to be ignored is not independently decided, which reduces the modeling flexibility in frame selection.

As the second approach we propose the \textit{Mask-based} model, where a frame-wise mask on top of the encoder\footnote{Here ``frame-wise'' actually means \emph{frame-wise after down-sampling}, in accordance with the frame down-sampling in the encoder network (see Section~\ref{sec:exp_setting} for details).} is estimated by leveraging the speaker acoustic information contained in the anchor word and the actual recognition utterance.  The attention mechanism is then performed on the masked feature representation. Compared with the Multi-source Attention model, attention in the Mask-based model only focuses on remaining frames after masking, and for each frame it is independently decided whether to be masked out based on their acoustic similarity. Formally, Eqn.~(\ref{eqn:bahdanau}) and Eqn.~(\ref{eqn:context}) are modified as:
\begin{eqnarray}
\phi_t &=& \mathrm{sigmoid}(g\cdot \mathrm{Similarity}(\mathbf{u}_t,\tilde{\mathbf{w}})) \label{eqn:sim2}\\
\mathbf{h}_t^{\textrm{masked}}&=&\phi_t \mathbf{h}_t \\
\omega_{n,t}&=&\mathbf{v}^\top \tanh(\mathbf{W}^q\mathbf{q}_n+\mathbf{W}^h \mathbf{h}_t^{\textrm{masked}}+\mathbf{b}) \\
\mathbf{c}_n&=&\sum_{t} \alpha_{n,t}\mathbf{h}_t^{\textrm{masked}}
\end{eqnarray}
where $\mathrm{Similarity}(\cdot,\cdot)$ in Eqn.~(\ref{eqn:sim2}) is dot-product as well.

\section{Synthetic Data and Multi-task Training}
\label{sec:synthetic}
\subsection{Synthetic Data}
A problem we encountered in our task is: there is very little training data that has the same condition as the test case. Some utterances in the test set contain speech from two or more speakers (denoted as the ``speaker change'' case), and
some of the other utterances only contain background speech (denoted as the ``no desired speaker'' case). In contrast, most of the training data does not have interfering or background speech, making the model unable to learn to ignore.

\begin{figure}[ht]
  \centering
  \includegraphics[width=0.39\textwidth]{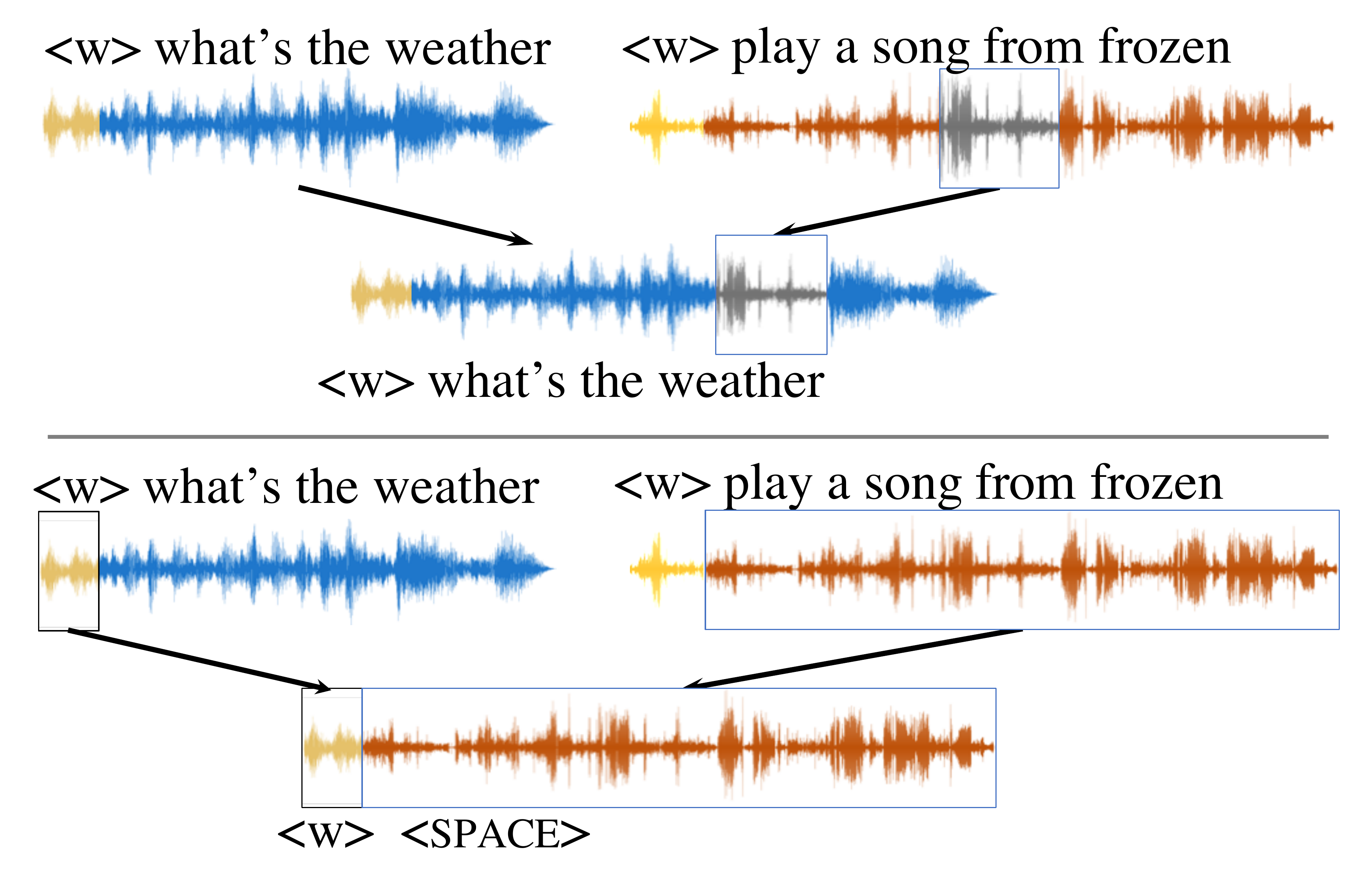}
  \vspace{-0.2cm}
  \caption{Two types of synthetic data: Synthetic Method~1 (top) and 2 (bottom). The symbol $\langle \textsc{w}\rangle$ represents the wake-up word, and $\langle\textsc{space}\rangle$ represents empty transcripts.}
  \label{fig:synthetic}
  \vspace{-0.25cm}
\end{figure}

In order to simulate the condition of the test case, we generate two types of synthetic data for training:
\begin{itemize}
\item Synthetic Method 1: for an utterance, a random segment\footnote{The frame length of a segment is uniformly sampled within the range [50,150] in our experiments. It is possible that the randomly selected segment is purely non-speech or even silence.} from another utterance in the dataset is inserted at a random position after the wake-up word part within this utterance, while its transcript is unchanged.
\item Synthetic Method 2: the entire utterance, excluding the wake-up word part, is replaced by another utterance, and its transcript is considered as empty.
\end{itemize}
Fig.~\ref{fig:synthetic} illustrates the synthesizing process. These two types of synthetic data simulate the ``speaker change'' case and the ``no desired speaker'' case respectively. The synthetic and device-directed data are mixed together to form our training data. The mixing proportion is determined from experiments.

\subsection{Multi-task Training for Mask-based Model}
\label{sec:multitask}
For the generated synthetic data, we know which frames come from the original utterance and which are not, i.e., we have the gold mask for each synthetic utterance, where the frames from the original utterance are labeled with ``1'', and the other frames are labeled with ``0''. Using this gold mask as an auxiliary target, we train the Mask-based model in a multi-task way, where the overall loss is defined as a linear interpolation of the normal ASR cross-entropy loss and the cross-entropy-based mask loss: $(1-\lambda)\mathcal{L}_\mathbf{ASR}+\lambda \mathcal{L}_\mathbf{mask}$.

The gold mask provides a supervision signal to explicitly guide $\mathrm{S{\text -}Encoder}$ to extract acoustic features that can better distinguish the inserted frames from those in the original utterance. As will be shown in our experiments, with the multi-task training the predicted mask is more accurate in selecting desired frames for the decoder.


\section{Experiments}
\label{sec:exp}
\subsection{Experimental Settings}
\label{sec:exp_setting}
\vspace{-0.5mm}
We conduct our experiments on training data of 1200-hour live data in English collected from the Amazon Echo. Each utterance is hand-transcribed and begins with the same wake-up word whose alignment with time is provided by end-point detection~\cite{shin2000speech,li2007robust,shannon2017improved,maas2018combining}. As we have mentioned, while the training data is relatively clean and usually only contains device-directed speech, the test data is more challenging and under mismatched conditions with training data: it may be noisy, may contain background speech\footnote{background speech includes: 1) interfering speech from an actual non-device-directed speaker; and 2) multi-media speech, meaning that a television, radio, or other media device is playing back speech in the background.}, or may even contain no device-directed speech at all. In order to evaluate the performance on both the matched and mismatched cases, two test sets are formed: a ``normal set'' (25k words in transcripts) where utterances have a similar condition as those in the training set, and a ``hard set'' (5.4k words in transcripts) containing the challenging utterances with interfering background speech. Note that both of the two test sets are real data without any synthesis. We also prepare a development set (``normal''+``hard'') with a similar size as the test sets for hyper-parameter tuning. 
For all the experiments, 64-dimensional log filterbank energy (LFBE) features are extracted every 10ms with a window size of 25ms. The end-to-end systems are grapheme-based and the vocabulary size is 36, which is determined by thresholding on the minimum number of character counts from the training transcripts. Our implementation is based on the open-sourced toolkit \textsc{OpenSeq2Seq}~\cite{kuchaiev2018openseq2seq}.

Our baseline end-to-end model does not consider anchor words. Its encoder consists of three convolution layers resulting in 2x frame down-sampling and 8x frequency down-sampling, followed by 3 Bi-directional LSTM~\cite{hochreiter1997long} layers with 320 hidden units. Its decoder consists of 3 unidirectional-LSTM layers with 320 hidden units. The attention function is \emph{Bahdanau Attention}~\cite{bahdanau2014neural}. The cross-entropy loss on characters is optimized using Adam~\cite{kingma2014adam}, with an initial learning rate 0.0008 which is then adjusted by exponential decay. A beam search with beam size 15 is adopted for decoding. The above setting is also used in our proposed models.

\vspace{-0.2cm}
\subsection{Multi-source Attention Model vs. Baseline}
\vspace{-0.1cm}
$\mathrm{S{\text -}Encoder}$ consists of three convolution layers with the same architecture as that in the baseline's encoder.

First of all, we compare Multi-source Attention Model and the baseline trained on the device-directed-only data, i.e., without any synthetic data. The results are shown in Table~\ref{tab:multi-source_baseline_original}.
\vspace{-0.4cm}
\begin{table}[ht]
  \caption{Multi-source Attention Model vs. Baseline with device-directed-only training data. The WER, substitution, insertion and deletion values are all normalized by the baseline WER on the ``normal'' set\protect\footnotemark. The normalization applies to all the tables throughout this paper in the same way.}
  \center
  \setlength\tabcolsep{2pt}
    \footnotesize
    \begin{tabular}{  c c c c c c c c }
    \toprule
    Model & Training Set & Test Set & WER & sub & ins & del & WERR(\%) \\
    \midrule
    \multirow{2}{*}{Baseline} & \multirow{2}{*}{\makecell{Device-\\directed-only}} & normal & 1.000 & 0.715 & 0.108 & 0.177 & --- \\
    \cmidrule{3-8}
                              & & hard & 3.354 & 1.762 & 1.123 & 0.469 & --- \\
    \midrule
    \multirow{2}{*}{\makecell{Mul-src. \\Attn.}} & \multirow{2}{*}{\makecell{Device-\\directed-only}} & normal & 1.015 & 0.731 & 0.115 & 0.169 & -1.5 \\
    \cmidrule{3-8}
                              & & hard & 3.262 & 1.746 & 1.062 & 0.454 & +2.8 \\
    \bottomrule
    \end{tabular}
  \label{tab:multi-source_baseline_original}
  \vspace{-0.3cm}
\end{table}
\footnotetext{For example, if WER for the baseline is $5.0\%$ for the ``normal'' set, and $25.0\%$ for the ``hard'' set, then the normalized values would be $1.000$ and $5.000$ respectively.}
\\The relative WER reduction (WERR) of Multi-source Attention on the ``hard'' set is 2.8\% and it is mostly due to a reduction in insertion errors. We also observe a slight WER degradation of 1.5\% relative on the ``normal'' set. It implies that the proposed model is more robust to interfering background speech.

Next, we further validate the effectiveness of the Multi-source Attention model by showing how synthetic training data has different impact on it and the baseline model respectively. Synthetic training data is prepared such that $50\%$ of the utterances in the training set are kept unchanged, $44\%$ are processed with Synthetic Method 1, and $6\%$ are processed with Synthetic Method 2. The ratio is tuned on the development set. This new training data is referred as ``augmented''  in all result tables. Table~\ref{tab:sythetic_original} exhibits the results. For the baseline model, the performance degrades drastically when trained on augmented data: the deletion errors on both of the ``normal'' and ``hard'' test sets get much higher. This is expected since without the anchor word the model has no extra acoustic information of which part of the utterance is desired, so that it tends to ignore frames regardless of whether they are actually from device-directed speech. On the contrary, for the Multi-source Attention model the WERR (augmented vs. device-directly-only) on the ``hard'' set is $12.5\%$, and WER on the ``normal'' set does not get worse. Moreover, the insertion errors on both test sets get reduced while the deletion errors increase much less than those in the case of the baseline model, indicating that by incorporating the anchor word information the proposed model effectively improves the ability of focusing on device-directed speech and ignoring others. This series of experiments also reveals significant benefits from using the synthetic data with the proposed model. In total, the combination of the Multi-source Attention model and augmented training data achieves $14.9\%$ WERR on the ``hard'' set, with only $1.5\%$ degradation on the ``normal'' set.
\begin{table}[ht]
  \vspace{-0.3cm}
  \caption{Augmented vs. Device-directed-only training data. Results with ``Device-directed-only'' training set are from Table~\ref{tab:multi-source_baseline_original} for clearer comparisons.}
  \begin{center}
  \setlength\tabcolsep{2pt}
    \footnotesize
    \begin{tabular}{  c c c c c c c c }
    \toprule
    Model & Training Set & Test Set & WER & sub & ins & del & WERR(\%) \\
    \midrule
    \multirow{4}{*}{Baseline} & \multirow{2}{*}{\makecell{Device-\\directed-only}} & normal &     1.000 & 0.715 & 0.108 & 0.177 & --- \\
    \cmidrule{3-8}
                           &   & hard & 3.354 & 1.762 & 1.123 & 0.469 & --- \\
    \cmidrule{2-8}
    & \multirow{2}{*}{Augmented} & normal & 3.215 & 1.223 & 0.038 & 1.954 & -221.5 \\
    \cmidrule{3-8}
                            &   & hard & 4.208 & 1.777 & 0.246 & 2.185 & -30.9 \\
    \midrule
    \multirow{4}{*}{\makecell{Mul-src. \\Attn.}} & \multirow{2}{*}{\makecell{Device-\\directed-only}} & normal & 1.015 & 0.731 & 0.115 & 0.169 & \textbf{-1.5} \\
    \cmidrule{3-8}
                           &   & hard & 3.262 & 1.746 & 1.062 & 0.454 & +2.8 \\
    \cmidrule{2-8}
    &\multirow{2}{*}{Augmented} & normal & 1.015 & 0.700 & 0.108 & 0.207 & \textbf{-1.5} \\
    \cmidrule{3-8}
                           &    & hard & 2.854 & 1.569 & 0.723 & 0.562 & $\textbf{+14.9}$ \\
    \bottomrule
    \end{tabular}
  \end{center}
  \label{tab:sythetic_original}
  \vspace{-0.5cm}
\end{table}

\subsection{Mask-based Model}
In the Mask-based model experiments, 3 convolution and 1 Bi-directional LSTM layers are used as $\mathrm{S{\text -}Encoder}$, as we observed that it empirically performs better than convolution-only layers. 
Due to the importance of using the augmented data for training our previous model, the same synthetic approach is directly applied to train the Mask-based model. Also, as we mentioned in Sec~\ref{sec:multitask}, multi-task training can be conducted since we know the gold mask for each synthesized utterance. Given the imbalanced mask labels, i.e., frames with label ``1'' (corresponding to those from the original utterance) constitute the majority compared with frames with label ``0'' (corresponding to those from another random utterance), we use weighted cross entropy loss for the auxiliary mask learning task, where the weight on frames with label ``1'' is $0.6$ and on those with label ``0'' is $1.0$, to counteract the label imbalance.

We first set the multi-task loss weighting factor $\lambda=1.0$ so that only the mask learning is performed. It turns out that around $70\%$ of frames with label ``0'' and $98\%$ with label ``1'' are recalled on a held-out set synthesized the same way as the training data, which demonstrates the effectiveness of estimating masks from the synthetic data. 

Then we perform ASR using the Mask-based model with and without mask supervision respectively, and the results are presented in Table~\ref{tab:mask}. WERRs are all relative to the baseline model trained on device-directed-only data. For the Mask-based model without mask supervision, it achieves $3.9\%$ WERR on the ``hard'' set while has a degradation of $34.8\%$ on the ``normal'' set. On the other hand, with mask supervision ($\lambda=0.1$) corresponding to the multi-task training, it yields $12.6\%$ WERR on the ``hard'' set while only $3.0\%$ worse on the ``normal'' set. The performance gap between them can be attributed to the ability of mask prediction: while with mask supervision the recall is still around $70\%$ (for frames labeled as ``0'') and $98\%$ (for frames labeled as ``1'') on the held-out set, it is only $48\%$ and $50\%$ respectively without mask supervision.

Note that even with multi-task training, the WER performance of the Mask-based model is still slightly behind the Multi-source Attention model, mainly due to the insertion error. Our conjecture is, the mask prediction is only done within the encoder, which may lose semantic information from the decoder that is potentially useful for discriminating device-directed speech from others.

\begin{table}[ht]
  \vspace{-0.3cm}
  \caption{Mask-based Model: with and without mask supervision.}
  \begin{center}
  \setlength\tabcolsep{2pt}
    \footnotesize
    \begin{tabular}{  c c c c c c c c }
    \toprule
    Model & Training Set & Test Set & WER & sub & ins & del & WERR(\%) \\
    \midrule
    \multirow{2}{*}{\makecell{w/o\\Supervision}} & \multirow{2}{*}{Augmented} & normal & 1.348 & 0.725 & 0.096 & 0.527 & -34.8\\
    \cmidrule{3-8}
                                 & & hard & 3.223 & 1.508 & 0.628 & 1.087 & +3.9\\
    \midrule
    \multirow{2}{*}{\makecell{w/\\Supervision}} & \multirow{2}{*}{Augmented} & normal & 1.030 & 0.715 & 0.115 & 0.200 &  \textbf{-3.0} \\
    \cmidrule{3-8}
                              & & hard & 2.931 & 1.586 & 0.809 & 0.536 & \textbf{+12.6} \\
    \bottomrule
    \end{tabular}
  \end{center}
  \label{tab:mask}
  \vspace{-0.5cm}
\end{table}

\vspace{-0.1cm}
\section{Conclusions and Future Work}
\vspace{-0.1cm}
\label{sec:conclusion}
In this paper we propose two approaches for end-to-end anchored speech recognition, namely \emph{Multi-source Attention} and the \emph{Mask-based} model. 
We also propose two ways to generate synthetic data for end-to-end model training to improve the performance. Given the synthetic training data, a multi-task training scheme for the Mask-based model is also proposed. With the information extracted from the anchor word, both of these methods show their ability in picking up device-directed part of speech and ignore other parts. This results in large WER improvement of 15\% relative on the test set with interfering background speech, with only a minor degradation of 1.5\% on clean speech. Obviously the mismatch still exists between the training and test data. Future work would include finding a better way to generate synthetic data with more similar condition to the ``hard'' test set, and taking decoder state into consideration when estimating the mask. The other direction is to utilize anchor word information in contextual speech recognition~\cite{chen2019end}.

\vspace{-0.15cm}
\section{Acknowledgements}
\vspace{-0.15cm}
The authors would like to thank Hainan Xu for proofreading.
\vfill

\bibliographystyle{IEEEbib}
\clearpage
\fontsize{8.8}{10.3}\selectfont
\bibliography{refs}

\end{document}